# INDUCTIVE INFERENCE AND THE REPRESENTATION OF UNCERTAINTY

N. C. DALKEY*

## 1. Introduction

The form and justification of inductive inference rules depend strongly on the representation of uncertainty. This paper examines one generic representation, namely, incomplete information. The notion can be formalized by presuming that the relevant probabilities in a decision problem are known only to the extent that they belong to a class K of probability distributions. The concept is a generalization of a frequent suggestion that uncertainty be represented by intervals or ranges on probabilities.

To make the representation useful for decisionmaking, an inductive rule can be formulated which determines, in a well-defined manner, a best approximation to the unknown probability, given the set K. In addition, the knowledge set notion entails a natural procedure for updating--modifying the set K given new evidence. Several non-intuitive consequences of updating emphasize the differences between inference with complete and inference with incomplete information.

## 2. Knowledge Sets

The basic frame of reference is an algebra of sets E; i.e., a set of events closed under disjunction and negation. In a medical context, for example, E might be the combined set of disease states, symptoms, and test results. It is assumed that there is a probability function $\underline{P}$ on E; however, $\underline{P}$ is not completely known. What is known is that $\underline{P}$ is contained in a set K of probability functions on E. Thus, some bounds on the probabilities may be known, or the expectation of one or more random variables may be known, and the like. In the case of complete ignorance, K is the class Z of all possible probability functions on E. For the case of complete information, K is a unit set. In the general case, K is some subset of Z.

Roughly, the size of the knowledge set K represents the degree of uncertainty concerning P. A large K indicates high uncertainty, a small K indicates fairly complete information. The relationship $K \subset K'$ imposes a partial order, in fact a lattice, on the subsets of Z. $K \subset K'$ can be interpreted as "K is less uncertain than K'". For a given event e, the interval max P(e) - min P(e) is a rough measure of the uncertainty concerning e.

---

*Computer Sciences Department, University of California, Los Angeles, Ca, 90024. This work was supported in part by National Science Foundation Grant IST 84-05161.

The knowledge set representation of uncertainty is the starting point for a number of approaches to decisions with incomplete information, including the game-against-nature [Wald, 50; Blackwell and Girshick, 54], several variants of maximum entropy methods [Jaynes, 68; Kullback, 59], and the author's min-score theory of induction, which is a generalization of the first two [Dalkey, 82].

One approach to uncertainty that has aroused considerable interest as a potential "inference engine" for expert systems is the Dempster-Shafer belief-function theory. [Shafer, 76; Gordon and Shortliffe, 84] At first glance, belief functions do not appear to fit the knowledge set representation; however, Kyburg has recently shown that that a belief function Bel can be mapped onto a class of probability functions K(Bel), namely, the set of P's such that P(e) ≥ Bel(e) for all e in E. [Kyburg, 85] On this representation, Bel(e) = min P(e). As Kyburg points out,
                                K(Bel)
knowledge sets are much more general than belief functions; any Bel can be expressed by a K, but most K's cannot be expressed by a Bel.

## 3. Inductive Inference

A knowledge set K--especially if it is large--is not a very useful guide for action. Belief functions, for example have been criticised on the grounds that they do not have a clear role in decision analysis.

The game against nature, maximum entropy methods and min-score theory bridge the gap between a knowledge set and decisions by invoking a form of inductive inference. By assumption, the actual probability $\underline{P}$ is in K; a natural question is, can a "best guess" be selected from K which is, in a reasonable sense, a best approximation to $\underline{P}$?

The question presupposes a measure of the goodness of the approximation. An appropriate measure is available in the theory of proper scores. Consider a partition of events E, and an estimate Q which is a probability distribution on E A score rule is a function S(Q,e) which assigns a rating to the estimate Q depending on which event e in E occurs. A score rule is called proper (admissable) if it fulfills the condition

$$\sum_E P(e)S(Q,e) \leq \sum_E P(e)S(P,e) \qquad (1)$$

That is, a score rule is called proper if the expectation of the score is a maximum when the estimate is the same

110

as the distribution which determines the expectation.

There is an infinite family of functions which fulfill (1). Among these are the logarithmic score $S(Q,e) = \log Q(e)$ and the quadratic score $S(Q,e) = 2Q(e) - \sum_E Q(e)^2$. An important sub-family consists of the decisional scores: Let $U(a,e)$ be the payoff if action a is taken and e is the state of nature. Let $a*(Q)$ be the optimal action assuming $Q$ is the distribution on the states of nature. $S(Q,e) = U(a*(Q),e)$ is a proper score. Decisional scores furnish an intimate tie between decision theory and the theory of inference.

It is convenient to introduce the definitions:
$G(P,Q) = \sum_E P(e)S(Q,e)$; $H(P) = G(P,P) = \sum_E P(e)S(P,e)$.

Given a knowledge set K, if a distribution Q is posited the actual expectation is $G(\underline{P},Q)$. The analyst would like to select a Q to make this quantity as large as possible; however, since $\underline{P}$ is unknown, a direct maximization of $G(\underline{P},Q)$ is undefinable. On the other hand, the analyst can determine the Q which maximizes the minimum (over P) of $G(P,Q)$. The maxmin clearly can be guaranteed and thus can be considered a lower bound to what can be obtained, knowing K.

An upper bound can be set by examining the value of additional information. In the theory of decisions with complete information, it is a theorem that additional information leads to a higher expectation. [Lavalle, 78] This result is often called the positive value of information (PVI) principle. There is no way to demonstrate PVI for incomplete information. However, the principle appears to apply a fortiori to the case of uncertainty, and hence is a reasonable candidate for a new postulate. In the previous section, it was noted that if $K \subset K'$, then K is more informative than $K'$. We thus are led to the postulate:

P1. If $K \subset K'$ and K is not empty, $V(K) \geqslant V(K')$

Here $V(K)$ designates the value of knowing K. An immediate consequence of P1 is $V(K) \leqslant V(P)$, P any member of K, or equivalently, $V(K) \leqslant \min_K V(P)$.

Despite the apparent weakness of P1, it has the strong consequence that a strict upper bound to the value of knowing K is the minimum value of any member of K. The expression $V(K)$ was introduced informally; however, it seems reasonable to interpret $V(P)$ as $H(P)$. Thus, we can summarize

$$\max_Q \min_P G(P,Q) \leqslant V(K) \leqslant \min_P H(P) \qquad (2)$$

From (1) we have $\min_P \max_Q G(P,Q) = \min_P G(P,P) = \min_P H(P)$.

111

Thus the bounds set by (2) are precisely the bounds set by the game against nature, where P is a strategy selected by nature, and Q is a strategy selected by the analyst. Many critics have objected to the game against nature on the grounds that it assumes "nature" is both rational and hostile; however, (2) does not involve either assumption. The upper bound is imposed by the requirement that additional information be of positive value.

If K is convex and closed, the game against nature has a solution. [Blackwell and Girshick, 54, theorem 2.5.1] For some scores such as the log score or the quadraditic score, where G(P,Q) is strictly concave in Q, there is a pure strategy for both the analyst and nature, and we can assert max min = min max--the lower and upper bounds coincide. In this case min H(P) is the unique solution, leading to the inference rule

> Min-score rule: Given a knowledge set K and a score rule S, select as a best guess the Q in K that minimizes H(Q).

For the logarithmic score, $H(P) = \sum_E P(e)\log P(e) =$ -Entropy(P). Thus, the min-score rule is equivalent to the maximum entropy rule for the log score. The min-score rule can also be applied to the associated class K(Bel) of a belief function Bel, thus affording a tie between belief functions and decision theory.

4. Updating

In addition to an inference rule, a complete theory of induction requires an updating procedure, i.e., a method of revising an estimate given new evidence. In effect, this entails a method of modifying the knowledge set K based on the new evidence, since the revised estimate can then be obtained by applying the min-score rule to the new K.

Unlike classic probability theory, updating with uncertainty requires two separate procedures, depending on whether the new evidence affects or does not affect the unknown probability P. Strictly, a knowledge class K should be envisaged as deriving from some body of evidence I, and thus be denoted by K(I), say. As long as K remains fixed within a problem, there is no need to formalize the dependence on I. If new evidence I' becomes available, however, the role of I' in modifying K must be made explicit.

One kind of new evidence does not change the underlying unknown probability P, but only what is known about P. As a simple example, suppose at first the investigator knows only the average of some random variable, but at a later date learns the variance as well. Clearly, the same P is involved;

all that has changed is that more information about $\underline{P}$ is now
available. We can call modifying K in light of evidence
which does not change the underlying probability <u>knowledge</u>
updating. If the appropriate knowledge set K(I') based
separately on the new evidence I' can be determined, then
clearly the new knowledge set K(I.I') = K(I).K(I')--the inter
section of the old and additional knowledge sets.

Knowledge updating is roughly analogous to Dempster's
orthogonal-sum composition for belief functions, but has a
number of advantages: (1) No assumptions concerning the
independence of the old and additional evidence is required.
(2) The procedure contains a built-in consistency check;
if K(I.I') is empty, then one or both of the two pieces of
evidence is incorrect. (3) The procedure fulfills the posi-
tive value of information in a strong way; K(I.I') ⊂ K(I),
and $\max_{K(I.I')} P(e) - \min_{K(I.I')} P(e) \leq \max_{K(I)} P(e) - \min_{K(I)} P(e)$ for every
event e.

The second type--<u>information</u> updating--extends classical
updating to the case of uncertainty. The proto-type is that
in which the new evidence I' consists of learning that some
event e in E has occurred. In this case, the new K consists
of conditioning each Q in the old K on the event e, or
formally

$K(I.e)$ = {P|there is a Q in K and $P = Q(\cdot|e)$}

It is noteworthy that since every Q in K is a complete proba-
bility function, everything is known to compute the condi-
tional function $Q(\cdot|e)$.

Informational updating is the analogue of the Shafer
conditional belief function Bel(A|B), but is considerably
more general, since it applies to K's which are not belief
functions, and even for belief functions may be more exten-
sive than the K generated by Bel(A|B).

No new consequences concerning min-score inference re-
sult from knowledge updating; however, direct application of
the min-score rule to K's derived from informational updat-
ing will generally lead to misestimates--a topic deserving
a separate section.

## 5. Min-score Inference with Information Updating

By definition, since $\underline{P}$ is in K(I), $\underline{P}(\cdot|e)$ is in K(I.e).
Thus, it would appear at first glance that the justification
of the min-score rule for K(I) would carry over to K(I.e).
However, problems can arise, illustrated by the following
example: Suppose we have an experiment with binary events
e and $\bar{e}$, and binary observations i and $\bar{i}$. Let p denote the
prior probability of e, q the likelihood of i given e, and

113

r the likelihood of i given $\bar{e}$. Suppose p is completely unknown, but q and r are known. Without loss of generality, we can suppose q > r. If we set p = r/(q-r), then the posterior of e given i is 1/2. Thus, for any symmetric score rule, the min-score estimate for the posterior is completely uninformative.

The difficulty is that the naive application of the min-score rule does not employ all the information that is available; it overlooks the fact that we know something about $\bar{i}$. To rectify this oversight, it is necessary to extend the analysis by treating the new K as an incompletely known information system. An information system consists of a set of events of interest or hypotheses E and a set of potential observations (data, signals, symptoms, messages, etc.) I. There is a joint probability distribution P(E.I) on hypotheses and observations; but the probabilities of direct interest are the conditional distributions P(E|I) of the events given the observations.

The expected score of an information system is defined as

$$H(E|I) = \sum_I P(i) \sum_E P(e|i) S(i,e)$$

where S(i,e) is shorthand for the score obtained if P(e|i) is the estimate and event e occurs. In words, the total expected score for the information system is the average of the expected scores for conditional estimates based on the observations.

For the case of uncertainty, where K is a set of joint distributions P(E.I), the appropriate application of the min-score rule is to select the joint distribution Q out of K which minimizes the expected score H(E|I). In that way, all of the information in K is taken into account. It can be shown that the two basic properties—guaranteed expectation and PVI—hold for this procedure. [Dalkey, 80]

As an elementary example, consider the case of the experiment with unknown prior cited above. K consists of all joint distributions on E and I derivable by setting p to any value between 0 and 1. If we opt for the logarithmic score, the inference consists in selecting out of K the minimally informative information system. An equivalent statement of the problem is to select out of the interval [0,1] a best-guess prior p* which minimizes H(E|I). The second formulation has some historical interest; the issue of dealing with unknown priors is is old as the theorem of Bayes. For this elementary example, p* can be found by solving for p the implicit equation

$$\frac{p}{1-p} e^{H(q)-H(r)} = \left(\frac{pq+(1-p)r}{p(1-q)+(1-p)(1-r)}\right)^{q-r} \quad (3)$$

(3) is not particularly "intuitive". In effect, it "downplays"--i.e., gives lower prior weight to--the more informative hypothesis.

If the example is extended to multiple observations, some surprising effects appear. The best-guess prior computed for a single observation is not the same as the prior computed for several observations; the prior is a function of the number of observations. This result suggests that the amount of information in large samples with unknown prior is decidedly less than indicated by classical sampling theory. In the case of complete information, a posterior distribution obtained from an observation can be used as a new prior to predict the occurrance of subsequent observations. This transfer is not valid for min-score priors; a new prior must be determined for the prediction. [Dalkey, 85] Thus, there is a basic difference between the diagnostic import and the prognostic import of observations, a fact of some consequence for the configuration of expert systems.